\documentclass[11pt]{article} 
\usepackage{rldmsubmit,palatino}
\usepackage{graphicx}
\usepackage{wrapfig}

\title{Functions \hspace{-0.30mm}that \hspace{-0.30mm}Emerge \hspace{-0.30mm}through \hspace{-0.30mm}End-to-End \hspace{-0.30mm}Reinforcement \hspace{-0.30mm}Learning\\
--- The Direction for Artificial General Intelligence ---}

\author{
Katsunari ~Shibata\thanks{http://shws.cc.oita-u.ac.jp/\~{}shibata/home.html} \\
Department of Innovative Engineering\\
Oita University\\
700 Dannoharu, Oita 870-1192, JAPAN \\
\texttt{katsunarishibata@gmail.com} \\
}

\begin{document}
\maketitle

\begin{abstract}
Recently, triggered by the impressive results in TV-games or game of Go
by Google DeepMind,
end-to-end reinforcement learning (RL) is collecting attentions.
Although little is known, the author's group has propounded this framework for around 20 years
and already has shown a variety of functions that emerge in a neural network (NN) through RL.
In this paper, they are introduced again at this timing.

``Function Modularization'' approach is deeply penetrated subconsciously.
The inputs and outputs for a learning system can be raw sensor signals and motor commands.
``State space'' or ``action space'' generally used in RL show the existence of functional modules.
That has limited reinforcement learning to learning only for the action-planning module.
In order to extend reinforcement learning to learning of the entire function on a huge degree
of freedom of a massively parallel learning system and to explain or develop human-like intelligence,
the author has believed that end-to-end RL from sensors to motors using a recurrent NN (RNN) becomes an essential key. Especially in the higher functions, since their inputs or outputs are difficult to decide, this approach is very effective by being free from the need to decide them.

The functions that emerge, we have confirmed, through RL using a NN
cover a broad range from real robot learning with raw camera pixel inputs
to acquisition of dynamic functions in a RNN.
Those are
(1)image recognition, (2)color constancy (optical illusion), 
(3)sensor motion (active recognition), (4)hand-eye coordination
and hand reaching movement, (5)explanation of brain activities, 
(6)communication, (7)knowledge transfer,
(8)memory, (9)selective attention, (10)prediction, (11)exploration.
The end-to-end RL enables the emergence of very
flexible comprehensive functions that consider many things in parallel
although it is difficult to give the boundary of each function clearly.
\end{abstract}

\keywords{\hspace{-6.5mm}
function emergence, end-to-end reinforcement learning (RL), recurrent neural network (RNN),
\\ \hspace{-5.8mm}higher functions, artificial general intelligence (AGI)
}

\acknowledgements{This research has been supported by JSPS KAKENHI Grant Numbers
JP07780305, JP08233204, JP13780295, JP15300064, JP19300070
and many our group members}

\startmain 

\section{Introduction}
Recently, triggered by the impressive results in TV-games\cite{DQN1,DQN2}
or game of Go\cite{AlphaGo} by Google DeepMind,
the ability of reinforcement learning (RL) using a neural network (NN)
and the importance of end-to-end RL is collecting attentions.
One remarkable point especially in the results in TV-games is the gap such that
even though the inputs of a deep NN are raw image pixels without any pre-processing
and the NN is just learned through RL,
the ability acquired through learning extends to the excellent strategies for several games.
The learning do not need special knowledge about learned tasks
and necessary functions emerge through learning,
and so it is strongly expected to open the way to the Artificial General Intelligence (AGI) or Strong AI.

It has been general that a NN is considered as just a non-linear function approximator for RL,
and a recurrent neural network (RNN) is used to avoid POMDP (Partially Observable Markov Decision Problem).
Under such circumstances, the origin of the end-to-end RL can be found in the Tesauro's work
called TD-gammon\cite{Tesauro}.
The author's group is the only one who has propounded this framework consistently for around 20 years
using sometimes the symbolic name of ``Direct-Vision-based Reinforcement Learning''\cite{ICNN97,Neurap98}
and has shown already a variety of functions that emerge in a NN or RNN through RL\cite{Intech}
although little is known about them unfortunately.

In this paper, the author's unwavering direction that end-to-end RL
becomes an important key for explaining human intelligence or developing human-like intelligence
especially for the higher functions is introduced at first.
It is also shown that a variety of functions emerge through end-to-end (oriented) RL;
from real robot learning with raw camera pixel inputs
to acquisition of dynamic functions in an RNN.
All of the works here have been published already, but
the author believes that it is worthwhile to know what functions emerge
and what functions hardly emerge at this timing
when the end-to-end RL begins to be focused on.

\section{The Direction for Human-like General Intelligence}
There is probably little doubt that human intelligence is realized thanks to the massively parallel
and cohesively flexible processing on a huge degree of freedom in our brain.
On the other hand, unlike in the case of unconsciousness,
our consciousness looks linguistic and so it is not parallel but sequential.
Therefore, it is impossible to completely understand the brain functions through our consciousness.
Nevertheless, the researchers have tried to understand the brain or develop
human-like intelligence by hands.
We are likely to divide a difficult problem into sub-problems
expecting each divided one is easier to solve.
Then ``Function Modularization'' approach has been deeply penetrated subconsciously.
However, for each module, we have to decide what are the inputs and outputs at first.
It is easily known that to decide the information that comes and goes between divided modules,
it is necessary to understand the entire function.
Then to decide the information, some simple frame is set in advance 
and that causes the ``Frame Problem''.
It can be also thought that the division into ``symbolic (logical) processing'' and ``pattern processing''
causes the ``Symbol Grounding Problem''.
In our brain, they may be processed in different areas, but
they must be closely related in the same brain as one NN.

``State space'' or ``action space'', which is generally used in RL,
can be another aspect of the function modularization.
Researchers have limited the learning only to the actions that make the mapping from state space
to action space, and have tried to develop the way of construction of state space
from sensor signals and given the way of generating motor commands from each action 
separately from RL.
It has also been taken for granted that in recognition problems, classification categories are given
and in control problems, reference trajectories are given in advance usually.
However, we can recognize complicated situations from a picture, but it is clear that
all of them cannot be given as classification targets in advance.
There is no evidence that reference trajectory is explicitly generated in our brain.
From the viewpoint of understanding and developing human-like general intelligence,
they are by-products of the function modularization.
They give unnecessary constraints on a huge degree of freedom,
and disturb the flexible and comprehensive learning despite the intension of the researchers.

Based on the above,
the author has thought that the interruption of human designers should be cut off
and the development of functions should be committed to learning of high-dimensional parallel systems
as much as possible. 
The inputs for a learning system can be raw sensor signals, the outputs can be
raw actuator signals, and the entire process from sensors to actuators (motors)
should be learned without dividing into functional modules.
A NN has an ability to optimize the parallel process
according to some value or cost function through learning.
If training signals are given by humans, they can be constraints for learning.
However, RL has an ability to learn autonomously
without receiving training signals directly
based on trials and errors using a motion-perception
feedback loop with the environment.

For the higher functions, which are different from the recognition or control,
it is difficult to decide either inputs or outputs, and that has disturbed the progress
of the research on them.
The author is expecting that higher functions also emerge through comprehensive
end-to-end RL from sensors to motors using a recurrent NN.
In image recognition, the use of deep learning makes the performance better
than the conventional approaches that need to design a feature space\cite{DeepImage}.
In speech recognition, to leave the learning to an RNN
makes the performance better than the combination with conventional methods like HMM\cite{DeepSpeech2}.
They seem to support the importance of end-to-end learning approach.

The author has thought that what emerges should be called ``functions'' rather than ``representation''
because it is not just a choice of representation,
but the result of the processing to get there.
Furthermore to be more accurate, the function that emerges cannot be localized clearly in the NN,
but it is just a label for us to understand through our consciousness.

\section{Functions that Emerge through Reinforcement Learning (RL)}
\vspace{-2mm}
Here, functions that have been observed to emerge in a NN through RL in our works are introduced.
The NN is trained using the training signals produced automatically by RL;
Q-learning, actor-critic or Actor-Q (for learning of both continuous motion and discrete action) based on TD learning.
By using actor outputs in actor-critic or actor-Q directly as motion commands,
not probability called policy but continuous motions have been learned directly.
The structure like convolutional NN is not used except for the work in \cite{Prediction}. 
The details of each work can be found in each reference.\vspace{-2mm}
\subsection{Static Functions that Emerge in a Layered Neural Network (NN)}
\begin{wrapfigure}[17]{r}{75mm}
\vspace*{-\intextsep}
\vspace{-8mm}
\centering
\includegraphics[height=4.8cm]{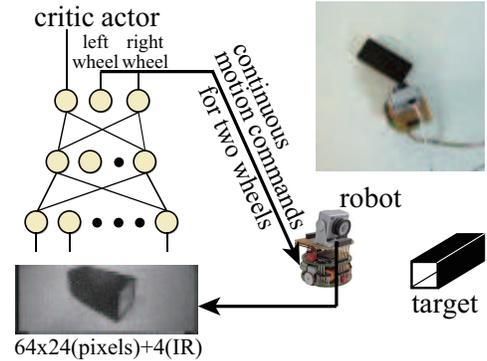}
\caption{Learning of Box Pushing: End-to-end reinforcement learning (RL) using a real robot (Khepera).
A reward was given only when the robot pushed the box, and continuous motions for the two wheels were learned. (2003)\cite{Khepera}}
\label{fig:BoxPushing}
\end{wrapfigure}
\vspace{-2mm}
A layered neural network (NN) is used and trained according to error back-propagation (BP),
and static functions emerge as follows.

\begin{figure}[b]
\centering
\vspace{-2mm}
\includegraphics[height=4.8cm]{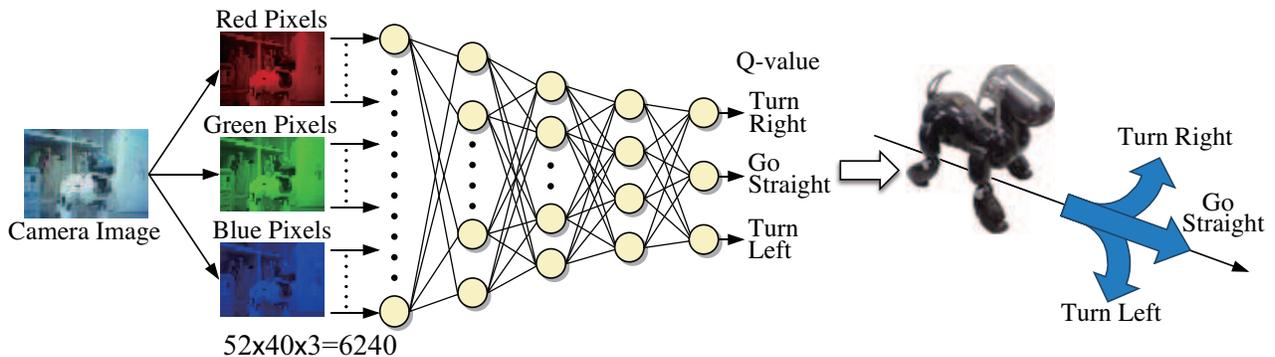}
\caption{Learning of approaching to kiss the other AIBO (not moving):
a real robot (AIBO) learned using a DQN (Deep-Q Network) (convolutional structure was not used)
in a real-world-like environment. (2008)\cite{AIBO}}
\label{fig:KissingAIBO}
\end{figure}
\noindent {\bf $\S$ Image Recognition}\\
Same as \cite{DQN1,DQN2}, images (pixels) were put into a neural network directly as inputs,
and appropriate behaviors to get a reward were acquired\cite{ICNN97,Neurap98}.
That was confirmed also in real robot tasks; Box Pushing (continuous motion)\cite{Khepera}
 and Kissing AIBO (real-world-like environment)\cite{AIBO,JCMSI}
 as shown in Fig. \ref{fig:BoxPushing} and Fig. \ref{fig:KissingAIBO}.\vspace{1mm}\\
\noindent {\bf $\S$ Color Constancy (Optical Illusion)}\\
Motion control of a colored object to the goal location
decided by the object color was learned.
The top view image covered by a randomly-appearing colored filter was the input.
From the internal representation,
we tried to explain the optical illusion of color constancy\cite{Color}.\vspace{0.5mm}\\
\noindent {\bf $\S$ Sensor Motion (Active Recognition)}\\
A pattern image was the input, and from the reward that indicates whether the
recognition result is correct or not,
both recognition and camera motion for better recognition
were acquired through RL\cite{ActiveP1}.
\vspace{0.5mm}\\
\noindent {\bf $\S$ Hand-Eye Coordination and Hand Reaching Movement}\\
A NN whose inputs were joint angles of an arm and also image on which its hand can be seen, and
whose outputs were the joint torques
learned to reach the hand to the randomly-located target that could be also seen on the image.
No explicit reference trajectory was given.
Furthermore, adaptation of force field and its after effect were observed\cite{Reaching}.\vspace{0.5mm}\\
\noindent {\bf $\S$ Explanation of the Brain Activations during Tool Use}\\
From the internal representation after learning of a reaching task with variable link length,
we tried to explain the emergence of the activities observed in the monkey brain
when the monkey used a tool to get a food\cite{Tool}.\vspace{0.5mm}\\
\noindent {\bf $\S$ Knowledge Transfer between Different Sensors}\\
An agent has two kinds of sensors and two kinds of motors.
There are four sensor-and-motor combinations.
There is a task that could be achieved using either of the combinations.
After the agent learned the task using 3 combinations,
learning for the remainder sensor-and-motor combination
was drastically accelerated\cite{KnowledgeTransfer}.\vspace{0.5mm}\\
\noindent {\bf $\S$ Game Strategy} (Not our works, but wonderful results can be seen
in \cite{Tesauro, DQN1, DQN2, AlphaGo})\\
\vspace{-1mm}

\subsection{Dynamic Functions that Emerge in a Recurrent Neural Network (RNN)}
Here, the emergence of dynamic functions is introduced in which expansion along the time axis is required.
In this case, an RNN is used to deal with dynamics,
and is trained by BPTT(Back Propagation Through Time) with the training signals generated automatically based on RL.
For both acquisition of memory and error propagation, the feedback connection weights are set
so that the transition matrix is the identity matrix or close to it when being linearly approximated.\vspace{0.5mm}\\
\noindent {\bf $\S$ Memory}\\
There are some works in which necessary information was extracted, memorized and reflected
to behaviors after learning almost only from a reward at each goal and punishment.
In \cite{Context}, a very interesting behavior in which if unexpected results occurred, an agent went back
to check the state in the previous stage without any direction could be observed.
In \cite{Meaning} and \cite{ActiveP2}, a real camera image was used as input,
and both camera motion and pattern meaning\cite{Meaning}
or both camera motion and word recognition\cite{ActiveP2} were learned.
Purposive associative memory could be also observed\cite{Meaning}.\vspace{0.5mm}\\
\noindent {\bf $\S$ Selective Attention}\\
It was learned that in a task in which the attended area of the next presented pattern is changed
according to the previously-presented image,
the area of the image was correctly classified without any special structure for attention\cite{Attention}.
Here, TD-based RL was not used, but learning was done by the reinforcement signal
representing only whether the final recognition result was correct or not.
Purposive associative memory could be also observed.\vspace{0.5mm}\\
\noindent {\bf $\S$ Prediction}
(shown in the next subsection\cite{Prediction})\vspace{0.5mm}\\
\noindent {\bf $\S$ Explanation of the Emergence of Reward Expectancy Neurons}\\
From a simulation result of RL using an RNN,
we tried to explain the emergence of reward expectancy neurons,
which responded only in the non-reward trials in a multi-trial task
observed in the monkey brain\cite{Ishii}.\vspace{0.5mm}\\
\noindent {\bf $\S$ Exploration}\\
Effective and deterministic exploration behavior for ambiguous or invisible goal considering past experience
was learned and temporal abstraction was discussed\cite{Exploration1,Exploration2}. \vspace{0.5mm}\\
\noindent {\bf $\S$ Communication} (introduced in another paper\cite{RLDM17COM})
\vspace{2mm}\\
Although each function has been examined in a very simple task, it is known that
a variety of functions emerge based on extraction and memory of necessary information
using an RNN.
However, the emergence of ``thinking'' or ``symbol processing'' that needs multi-step state transition
has not been observed yet.

\subsection{Frame-free Function Emergence for Artificial General Intelligence}
\begin{figure}[bh]
\center
\includegraphics[height=6.7cm]{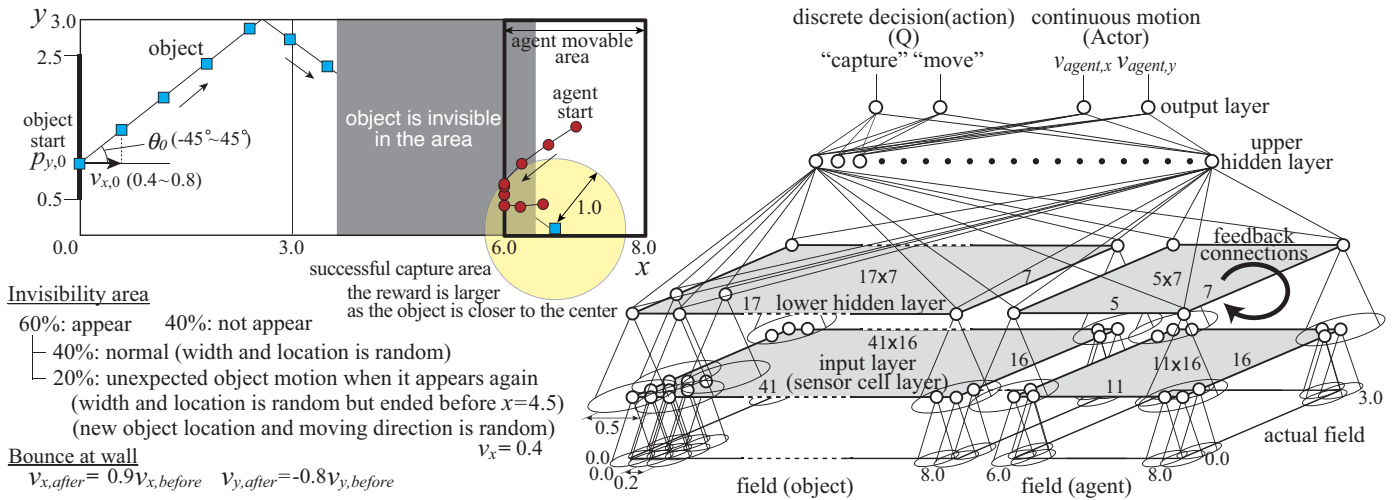}
\caption{Invisible moving object capture problem (prediction task). Invisibility area,
object start location, initial object moving direction and velocity are decided randomly
at each episode.
The input of the RNN is 832 visual signals (656 for the object and 176 for the agent itself).
By using Actor-Q, the agent can choose `capture' or `move', and when `move' is chosen,
the continuous motion is determined by the two actor outputs.(2013)\cite{Prediction}}
\label{fig:Predict}
\end{figure}

As mentioned before, through end-to-end RL from sensors to motors,
entire process is learned and comprehensive function is acquired.
For example, in the work in \cite{Prediction},
an agent who has a visual sensor and an RNN
as shown in Fig. \ref{fig:Predict} learned both motion and capture timing of a moving object
that often becomes invisible randomly.
\begin{wrapfigure}[17]{r}{80mm}
\centering
\includegraphics[height=6.3cm]{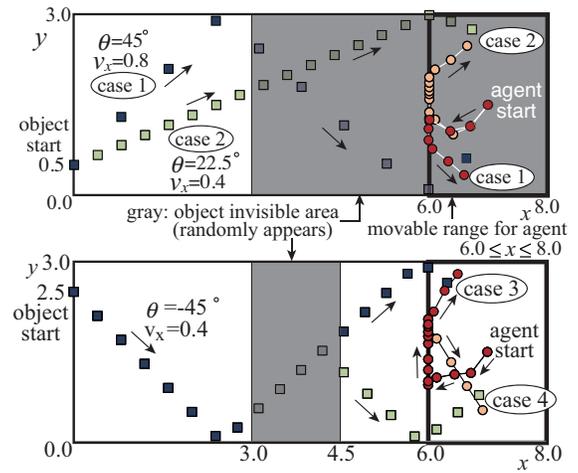}
\caption{Sample agent behaviors after learning of a prediction task. (2013)\cite{Prediction}}
\label{fig:PredictSamples}
\end{wrapfigure}
In a general approach, the object motion is estimated from some frames of image
using some given model, the future object location is predicted,
the capture point and time are decided by some optimization method,
a reference trajectory is derived from the capture point,
and the motions are controlled to follow the trajectory.

Fig. \ref{fig:PredictSamples} shows four examples after RL
based on the reward given for object capture.
The agent did not know in advance the way to predict the motion
or even the fact that prediction is necessary to catch it.
Nevertheless, it moved to the very front of its range of motion,
waited the object, and when the object came to close, the agent moved backward
with it and caught it.
Though the object became invisible or visible again suddenly,
the agent could behave appropriately.
Since the moving direction of the object changed sometimes
when it was invisible during learning, the agent learned to wait
close to the center ($y=1.5$) where it can react the unexpected
object motion.
As shown in case 4, when the object changed its direction unexpectedly,
the agent could catch it though the timing is a bit later
than the case of expected motion (case 3).


\fontsize{8.8}{0pt}\selectfont

\end{document}